\documentclass[runningheads]{llncs}
\usepackage[T1]{fontenc}
\usepackage{graphicx}
\usepackage{amsmath}
\usepackage{booktabs}
\usepackage{graphicx}
\usepackage{pgfplots}
\usepackage{tikz}
\usepackage{hyperref}
\pgfplotsset{compat=1.18, width=7.7cm}

\usepackage{subfig}
\usepackage{overpic}
\usepackage[misc]{ifsym}

\usepackage{microtype}
\begin{document}
\title{SPA: Towards A Computational Friendly Cloud-Base and On-Devices Collaboration Seq2seq Personalized Generation with Causal Inference}
%
%
\author{Yanming Liu\inst{1}\Letter \and
Xinyue Peng\inst{2}\Letter \and
Ningjing Sang\inst{3} \and
Yafeng Yan\inst{4} \and
Xiaolan Ke\inst{5} \and
Zhiting Zheng \and
Shaobo Liu \and
Songhang Deng \inst{6} \and
Jiannan Cao\inst{7} \and 
\\ Le Dai\inst{8} \and
Xingzu Liu \inst{9} \and
 Ruilin Nong \inst{9} \and
 Weihao Liu}
\authorrunning{Y. Liu et al. }
%
\institute{
Zhejiang University \\ \email{oceann24@zju.edu.cn} \and
Southeast University \\ \email{xinyuepeng@seu.edu.cn}  \and
The Fu Foundation School of Engineering and Applied Science, Columbia University \and
Stevens Institute of Technology \and 
Harvard University \and 
MIT \and 
UCLA \and
Beijing Institute of Technology \and
Tianjin University 
}
\maketitle                
\begin{abstract}
Large language models(LLMs) have shown its outperforming ability on various tasks and question answering. However, LLMs require substantial memory storage on low-resource devices. More critically, the computational speed on these devices is also severely limited. In this paper, we propose SPA(Side Plugin Adaption), a lightweight architecture for fast on-devices inference on the constraints of strict on-devices computation and memory constraints. Compared with other on-devices seq2seq generation, SPA could make a fast and stable inference on low-resource constraints, allowing it to obtain cost effiency. Our method establish an interaction between a pretrained LLMs on-cloud and additive parameters on-devices, which could provide the knowledge on both pretrained LLMs and featured personal feature. Further more, SPA provides a framework to keep feature-base parameters on  low computational devices while leave the parameters containing general information on the high computational devices. Our code is public at \url{https://github.com/OceannTwT/spa}.

\keywords{Personalized LLM; Cloud-device Collaboration; Inference Acceleration. }
\end{abstract}
\section{Introduction}
Large Language Models (LLMs)\cite{brown2020language,touvron2023llama} have demonstrated incredibly powerful capabilities in supporting different systems\cite{bo2024attention,xinjin2024intell}. These models perform exceptionally well across various downstream tasks\cite{xu2024applications,gu2024predicting}. Additionally, different LLMs such as GPT-4\cite{openai2023gpt}, Llama\cite{touvron2023llama} exhibit similar abilities in handling tasks across multiple dimensions\cite{bang2023multitask}. However, deploying and computing LLMs on resource-constrained devices introduces significant memory and computational pressure on these devices\cite{alizadeh2023llm}. Therefore, researching reliable methods for deploying LLMs on-devices becomes critically urgent.

\begin{figure}
\centering
\resizebox{\linewidth}{!}{
    \includegraphics[width=0.6\linewidth]{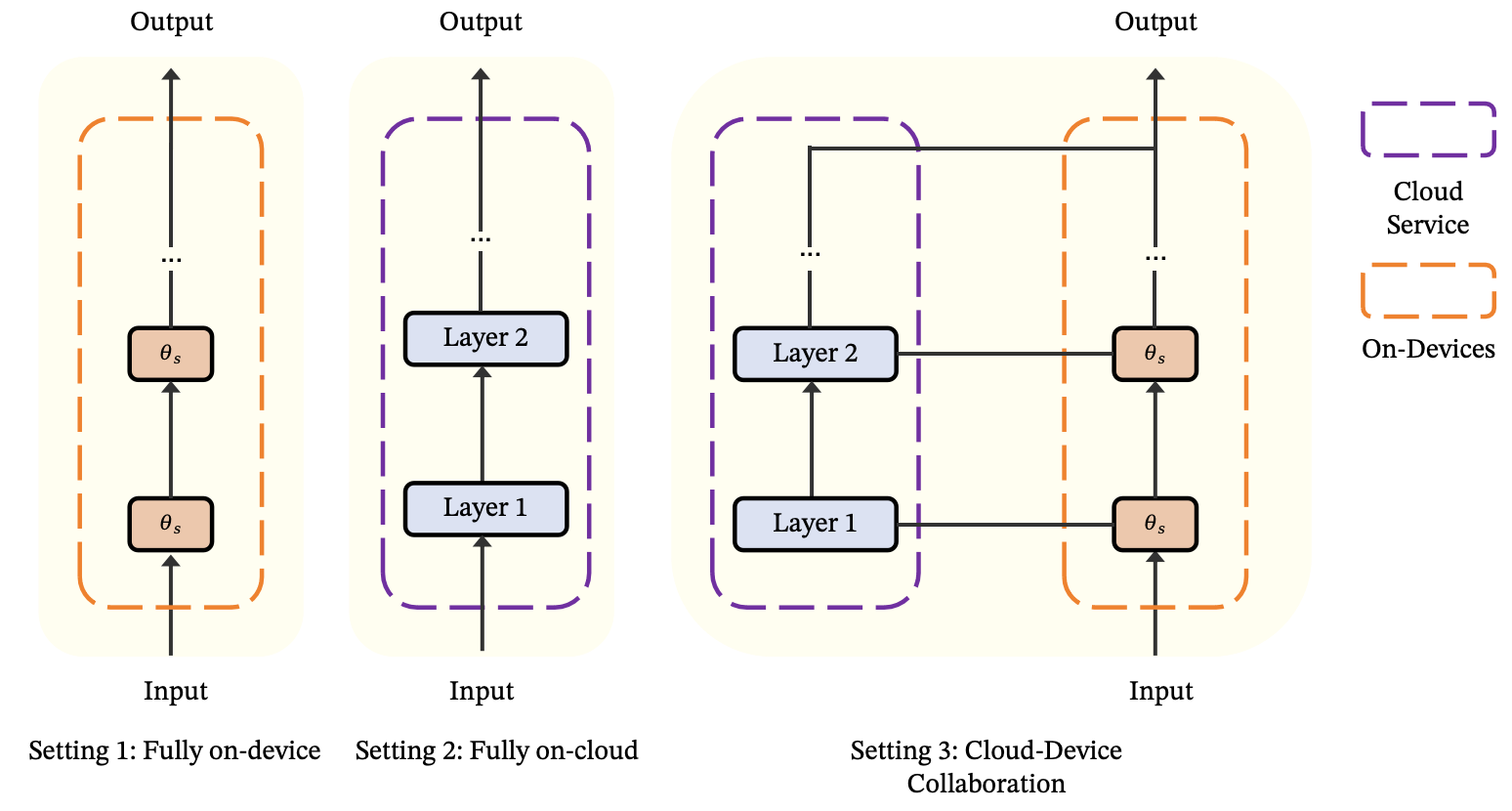}
    }
    \caption{
Three methods for model deployment are based on different combinations of cloud and device setups.}
    \label{fig:three-setting}
\end{figure}

To address the issue of deploying large language models (LLMs) on devices, various methods have been proposed to adapt them for edge devices. Subformer \cite{reid-etal-2021-subformer-exploring} reduces storage and computation requirements by sharing parameters. Similarly, EdgeFormer \cite{ge-etal-2022-edgeformer} proposes a balanced sharing scheme, exploring the impact of different architectures on device deployment. LLMCad \cite{xu2023llmcad} employs a collaborative model configuration to achieve lightweight deployment, where a small model generates simple words and a high-accuracy model verifies and corrects errors. This study builds on the principles of uncertainty quantification and sparse estimation of spectral lines\cite{10021978}, applying similar causal inference techniques to enhance personalized generation models in cloud-based and on-device collaborative frameworks.

Additionally, research has proposed parameter-efficient fine-tuning methods for LLMs \cite{houlsby2019parameter,hu2021lora}, utilizing adapter strategies to flexibly adapt to downstream tasks while preserving model capabilities. Adapters are stored separately from the base model and require only a few parameters to customize for downstream tasks\cite{hu2023llm}. For on-device deployment, adapters can be used for incremental deployment, integrating incremental knowledge into existing models to support downstream tasks and obtain user attributes\cite{lin2020exploring,zhong2024deep,gao2016novel,wang2024research,xinjin2024a}. However, this approach requires consideration of data transmission requirements between devices \cite{ren2019collaborative}, and reducing transmission latency is a significant challenge in the design of separated models\cite{yang2024research}.

Our research proposes a method called \textbf{S}ide \textbf{P}lugin \textbf{A}daptation (\textbf{SPA}), which achieves the deployment of large models on-devices by separating adapters from pre-trained models. Specifically, our method introduces an side adapter classifier. This classifier allows us to choose between leveraging the inherent capabilities of the original model and integrating feature information generated by adapters. In this way, we aim to minimize the frequency of adapter usage while maximizing the utilization of common-sense knowledge inherent in the original large language model. This strategy offers a dual advantage: leveraging the unique feature attributes of adapters while maximizing the inherent capabilities of a widely based model. We also evaluate the effect on casual inference. Compared to traditional on-devices model setups, our method significantly improves inference speed. 

We comprehensively evaluate our approach on various datasets across different domains. These evaluations not only emphasize the advantages of our method compared to traditional approaches but also highlight its potential in improving overall model capability while achieving significant adapter effects. In addition to enhancing text generation and semantic understanding services, our approach demonstrates a stronger ability to understand textual features, better meeting user needs.

Our contributions can be summarized as follows:

\begin{itemize}
\item[$\bullet$] 
We innovatively propose the SPA framework, deploying large models on-device via adapters. Our method significantly reduces the difficulty of fully deploying large models on the edge and presents a feasible and efficient solution. 

\item[$\bullet$] 
We introduce a classifier to determine whether to use adapter content for inference. This innovative classifier design significantly improves the efficiency of cloud server to edge transmission while maintaining task performance. 

\item[$\bullet$] 
SPA has been thoroughly evaluated across multiple datasets. Experiments show that our framework performs well on-devices. Additionally, SPA can achieve results close to the original model with lower transmission latency. 

\end{itemize}

\section{Related Work}

\subsection{Parameter-Efficient Transfer Learning}
In recent years, efficient parameter-efficient transfer learning has become a popular method for fine-tuning large language models. It achieves performance close to full fine-tuning by refining a subset of parameters. Among these methods, adapters, as introduced by \cite{houlsby2019parameter}, incorporate two additional feed-forward layers, enhancing their adaptability to downstream tasks. Similarly, LoRA \cite{hu2021lora} employs a rank reduction followed by rank promotion method on the original language model. In the context of language model fine-tuning, prefix tuning \cite{li2021prefix} involves introducing a trainable prefix vector before the keys and values in the multi-head attention mechanism, producing effects similar to the first two methods.

\subsection{On-device ML optimization}

For on-device sequence-to-sequence models, various techniques have been adopted to reduce the parameters of pre-trained models\cite{ma2024fostc3net}, including methods such as quantization\cite{li2023loftq} to reduce model size, knowledge distillation\cite{wang2021knowledge} techniques that learn from large models and transfer knowledge to smaller models, and techniques such as compression\cite{frantar2023gptq} and pruning\cite{bolya2022token} to reduce the number of model parameters. Some methods like Edgeformer \cite{reid2021subformer} achieve model reduction by sharing model parameters. Low-rank decomposition is also used to reduce model size \cite{cahyawijaya2021greenformers}. These techniques are adjusted and adapted based on the parameters of the Transformer  architecture, and they yield effective parameter reduction results for models with relatively fewer parameters.

\section{Methodology}
\label{sec:spa}

Considering the edge computing and storage capabilities, as well as the need to learn user characteristics and contextual information related to text generation, we propose \textbf{SPA}, a lightweight architecture for fast on-devices inference on the constraints of strict on-devices computation and memory constraints. With this design, we can evaluate its support for downstream tasks and compare it with the results of the original edge adapter.

\subsection{Side Model Design}

Collaborative learning and training between edge and cloud models currently face the following major bottlenecks: while all models are deployed on edge devices, the computation required for the models involves extensive floating-point operations, and the memory needed for model training significantly increases. For the autoregressive model, these issues make it tough to directly use large language models (LLMs) for inference on edge devices\cite{10526478}. On the other hand, when data is placed in the cloud, although general computing power is ensured, some personalized features cannot be efficiently transferred from the edge to the cloud for effective learning. These bottlenecks highlight the urgent need to design a model that combines both edge and cloud capabilities to better adapt to the respective requirements.

To address this, we propose integrating causal inference techniques into the SPA framework. By leveraging causal inference, we can systematically identify the effect of different model settings (e.g., edge-only, cloud-only, or hybrid) on the performance of personalized text generation. Specifically, we place some learnable parameters on the edge to enable rapid personalization of information learning settings. SPA involves using a classifier to select the output results generated by the original LLM or the side block. The classifier optimizes the decision-making process by estimating the causal effect of choosing different paths, enhancing the overall quality of text generation. Additionally, the classifier can decide whether to combine the edge model with the base model stored in the cloud. In this way, the edge model only needs to learn the feature differences relevant to the current task\cite{tao2023mlad}, while the general information is provided by the base model, reducing the resource latency costs on edge devices.

Specifically, if we define function \( f \) as the single layer on the original LLM and function \( g \) as the corresponding layer on the side model, the causal effect of choosing the output can be formalized using the following functions:

\[
\mathcal{H}_i(x) = f_i(x) + \sigma_i g_i(x)
\]

Here, \( \sigma_i \) represents the decision variable that chooses between the outputs of \( f \) and \( g \). This can be viewed as estimating the conditional average treatment effect (CATE) of selecting a path based on the input features:

\[
CATE(x) = E[\mathcal{H}_i(x) | \text{Select } g] - E[\mathcal{H}_i(x) | \text{Select } f]
\]

By using a linear layer of \( H \times 2 \) as our classifier, the model predicts the optimal path based on this CATE estimation, leveraging the Transformer's hidden state. The classifier output is softmaxed to range between \( [0, 1] \), allowing for a probabilistic selection given by:

\[
\sigma_i = \arg\max(\textbf{M}_{c}f_i(x))
\]

The model learns whether predicting a certain token is better accomplished by selecting the existing model’s output or by employing the side-end incremental model that relies on user-specific feature information. This choice hinges on the estimated causal effect, ensuring that the model dynamically adapts to the task requirements.\cite{bo2024rootcauseattributiondelivery,xiao2024multiple} Since \( \sigma_i \) is learnable, the overall loss function can be expressed as:

\[
\mathcal{L} = \sum_i \left( \text{Loss}(f_i, y) + \text{CATE}(\sigma_i) \right)
\]

The loss function now not only accounts for prediction accuracy but also incorporates the causal impact of model decisions, driving the SPA model towards optimal, personalized performance.

\begin{figure*}[t]
    \centering
    \includegraphics[width=1\linewidth]{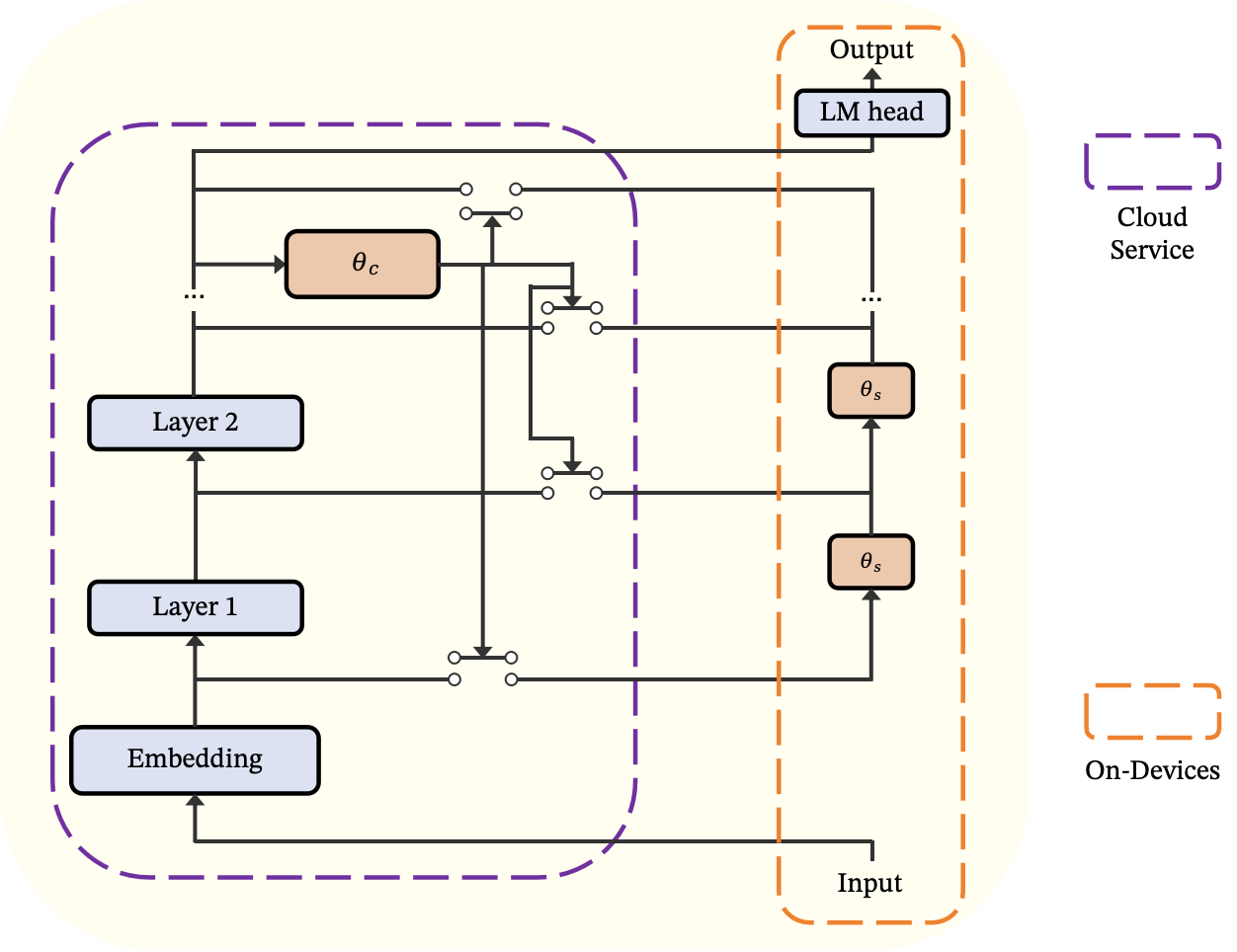}
    \caption{
Cloud and Device Collaboration Model SPA. This framework utilizes a classifier to provide optional downstream task computation choices for edge devices, thereby reducing computational pressure on edge devices and improving the overall performance of specialized device models.}
    \label{fig:spa}
\end{figure*}

\begin{equation}
    \text{Loss}(f_i, y) = - \sum_{i = 1}^T \log_{(\Theta+ \sigma \Delta \Theta)}{\textbf{P}(x_i|x_{<i})}
\end{equation}

Here, we keep \( f(x) \) frozen and unchanged, while training the side model \( g(x) \) and the classifier \( \sigma \). Unlike the fixed classifier \( \sigma' \) used for text similarity in the original model, our classifier \( \sigma \) based on a linear layer can determine whether the predicted word exists in the knowledge base of the underlying language model. This approach reduces the inference time for highly repetitive word sequences, stores more general information in the cloud, and generates user-specific features on edge devices. Compared to EdgeFormer, the main bottleneck of our model is the communication latency between the edge device and the server. This can be optimized by controlling the frequency and timing of communication. While our approach results in a smaller edge model, its prediction speed relative to the total latency is very fast. In this case, the computational power of the edge device will not become a bottleneck for the prediction results.

\subsection{Communications Latency}

For communication between cloud servers and edge models, we adopt the architecture used in LST. LST improves memory capacity through backpropagation. For prediction, if our transformer has $L$ layers, the Parallel adapter or LoRA method results in higher communication latency costs. Specifically, the latency includes three parts: $T_{pretrained}$ represents the computation latency of the original pre-trained model, $T_{on-devices}$ represents the computation latency of the edge device model, which could be the Adapter part, LoRA part, or LST part used for edge device inference. Additionally, $T_{net}$ is the network latency when transmitting data between the cloud and devices. Then, the total latency can be evaluated as:

\begin{equation}
    T_{total} = T_{on-devices} + T_{pretrained} + T_{net}
\end{equation}

While $T_{pretrained}$ keep the same for identical model. For model on edge devices, the inference latency is lay on the CPU or GPU computing capability, denoted as $f^e$. $F_{data}$ is the FLOPs required to compute on edges, $C_{devices}$ indicates the numbers of CPU or GPU are available to compute on this model. Then:

\begin{equation}
    T_{on-devices} = \frac{F_{data} \times {C_{devices} }}{f^e}
\end{equation}

For the network latency when transmitting data between cloud and devices. Assume the TCP/IP connection latency is $\tau$, The $T_{net}$ could be formulated as:

\begin{equation}
    T_{net} = \mathcal{M} \times (\tau + T_{data})
\end{equation}

Where $\mathcal{M}$ is the transmission times for one token prediction, it would be changed for different architecture, because we have to transfer back the hidden state from side model to base model\cite{tao2023sqba}. SPA effectively merges the process in scenarios where edge-cloud connectivity is involved.

\section{Experiments Setup}

\subsection{Datasets}

We evaluated our approach on the following training dataset. 

\begin{enumerate}
    \item \textbf{XSum}\cite{narayan2018don}: There are 227, 000 summaries of online articles generated by the BBC available on XSum, and we assessed the relationships between the articles and their corresponding summaries. 
    \item \textbf{CNN Daily Mail}\cite{chopra2016abstractive}: The CNN Daily Mail dataset consists of 312, 000 summaries of CNN Daily Mail articles and its origin articles. 
    \item \textbf{CoQA}\cite{reddy2019coqa}: CoQA datasets contains 127, 000 questions encompassing 8, 000 conversation-based question-answer pairs, constituting a large-scale QA system spanning various domains. 
    \item \textbf{SciQ}\cite{welbl2017crowdsourcing}: The SciQ dataset comprises 13, 679 publicly available scientific exam questions covering diverse disciplines such as biology, chemistry, physics, and others\cite{qian2020heterogeneous}. Answers to specific questions are obtained through contextual information. 
\end{enumerate}

\begin{table}[t]
\centering
\vspace{0.2cm}
\caption{Evaluations on side model ratios. }
\begin{tabular}{lllll}
\toprule
Data sizes           & $\text{SPA}_\text{small}$ & $\text{SPA}_\text{medium}$ & $\text{SPA}_\text{full}$  \\ \midrule
Usage Percentage & 82.3  & 71.5 & 61.8 \\
\bottomrule
\end{tabular}
\label{tab:777}
\end{table}

\subsection{Implement Details}

For each dataset, we uniformly trained for 15 epochs to enhance the adapter's learning capabilities for tasks or text features. Simultaneously, when evaluating our sequenced model, we conducted branch selection checks on the evaluated content to observe whether the model opted for adapters for adaptation. We also statistically analyzed the proportion of adapter usage. We compared our models under consistent trainable parameters, setting the learning rates to $\{$ ${2\times10^{-4}, 5\times10^{-4}, 1\times10^{-3}}$ $\}$ for each model to determine the optimal learning rate for that dataset. Throughout the training process, we maintained a consistent batch size of 8 and conducted training across an 8 Nvidia RTX V100 machines. Table~\ref{tab:main_result} shows the comparison between our proposed model and the baseline model, as well as the LST model, demonstrating the differences in performance across various datasets. We utilize beam search for tokens generation. 

\section{Experiments}

\subsection{Main Result}

\textbf{The performance of SPA surpasses that of LST. } Table~\ref{tab:main_result} shows  that, under the premise of lightweight parameters, the blocks trained by SPA can effectively guide the original model. Consequently, even with a smaller computational load at the edge, the model can exhibit strong inference capabilities and enhance performance for the given task.

\begin{table*}[t]
\caption{
\small
Latency evaluated from Different efficient tuning methods. Measuring new parameters percentage rate, Inference latency, Network latency and Total latency on prediction for 50 new tokens. \textbf{Ratios} indicates the ratios of direct transmission times and pretrained model inferencing times. The Rouge-L scores on different PEFL methods and our model on various datasets. 
}
\resizebox{\linewidth}{!}{
\begin{tabular}{llllllll}
\toprule
\textbf{Model} & \textbf{\% Param} &\textbf{Infer. latency} & \textbf{Net. latency} & \textbf{Total latency} & \textbf{Ratios} & \textbf{XSum} & \textbf{CoQA} \\
\midrule
LLaMA-7B + LoRA & 0.19 & \textbf{3.26}  & 6.37 & 9.63 & 32.0 & \textbf{38.24} & 40.32 \\
LLaMA-7B + Adapter & 0.38 & 3.32 & 12.56 & 15.88 & 64.0 & 38.18 & \textbf{40.51} \\
LLaMA-7B + LST & \textbf{0.2} & 3.29 & 0.31 & 3.60 & 1.0 & 28.78 & 31.24 \\
LLaMA-7B + SPA & 0.21 & 3.30 & \textbf{0.18} & \textbf{3.48} & \textbf{0.62} & 35.52 & 37.30 \\
\bottomrule
\end{tabular}}
\label{tab:222}
\end{table*}

In the SPA architecture, network transmission is a crucial aspect, particularly in the context of SPA. Therefore, as the discussion in Section~\ref{sec:spa}, it's essential to compare the previous model with the SPA model and estimate the corresponding latency required for predicting the entire model. Given that our model is relatively large, we cannot deploy the entire model at the edge as in the case of Edgeformer. In the collaborative process between cloud-based and on-device components, transmitting critical information is necessary. Consequently, we will be conducting inference latency experiments to assess this aspect. 

\subsection{Latency on Prediction}

To address concerns about network latency and overall inference delay, we conducted comparative experiments on data exchange for these models. Additionally, we compared the original Adapter, LoRA, and LST methods, evaluating them based on data exchange times and overall model latency. We performed these assessments using the same training configuration as mentioned above. 

We also uniformly evaluated the time taken to generate 50 tokens. Additionally, our single-sided model and the baseline model were placed within a server cluster to simulate network transmission delays in an ideal network environment. This was done to assess the corresponding network transmission latency. The respective metric results obtained from the testing are recorded in Table~\ref{tab:222}. the difference between single and multiple communication latency is substantial, with inference costs relying mainly on cloud server computing power. 

\begin{table}[t]
\centering
\vspace{0.2cm}
\caption{
The Rouge-L scores on different settings and SPA model on various datasets. 
}
\begin{tabular}{llllll}
\toprule
\textbf{Model} & \textbf{Ratios} & \textbf{XSum} & \textbf{CNN-DM} & \textbf{CoQA} & \textbf{SciQ}\\
\midrule
LLaMA-7B + Fully Integration & 1.0 & 36.24 & \textbf{40.55} & \textbf{38.81} & \textbf{25.91} \\
LLaMA-7B + Fully Separation & 0.68 & 30.29 & 34.19 & 32.11 & 23.71 \\
LLaMA-7B + LST & 1.0 & 28.18 & 32.15 & 31.24 & 23.24 \\
LLaMA-7B + SPA & \textbf{0.64} & \textbf{36.52} & 39.22 & 37.30 & 25.38 \\
\bottomrule
\end{tabular}
\label{tab:333}
\end{table}

\textbf{
The SPA model outperforms other methods in terms of network latency.} Notably, SPA has many excellent features. It can make predictions by choosing between the pre-trained model and the integrated adapter model through a classifier. Due to the use of the LST structure, SPA's architecture is more independent on devices, giving it an advantage over methods like LoRA and Adapters.

\begin{figure*}[t] 
  \begin{minipage}[b]{0.46\textwidth}
    \centering
    \includegraphics[width=0.98\linewidth]{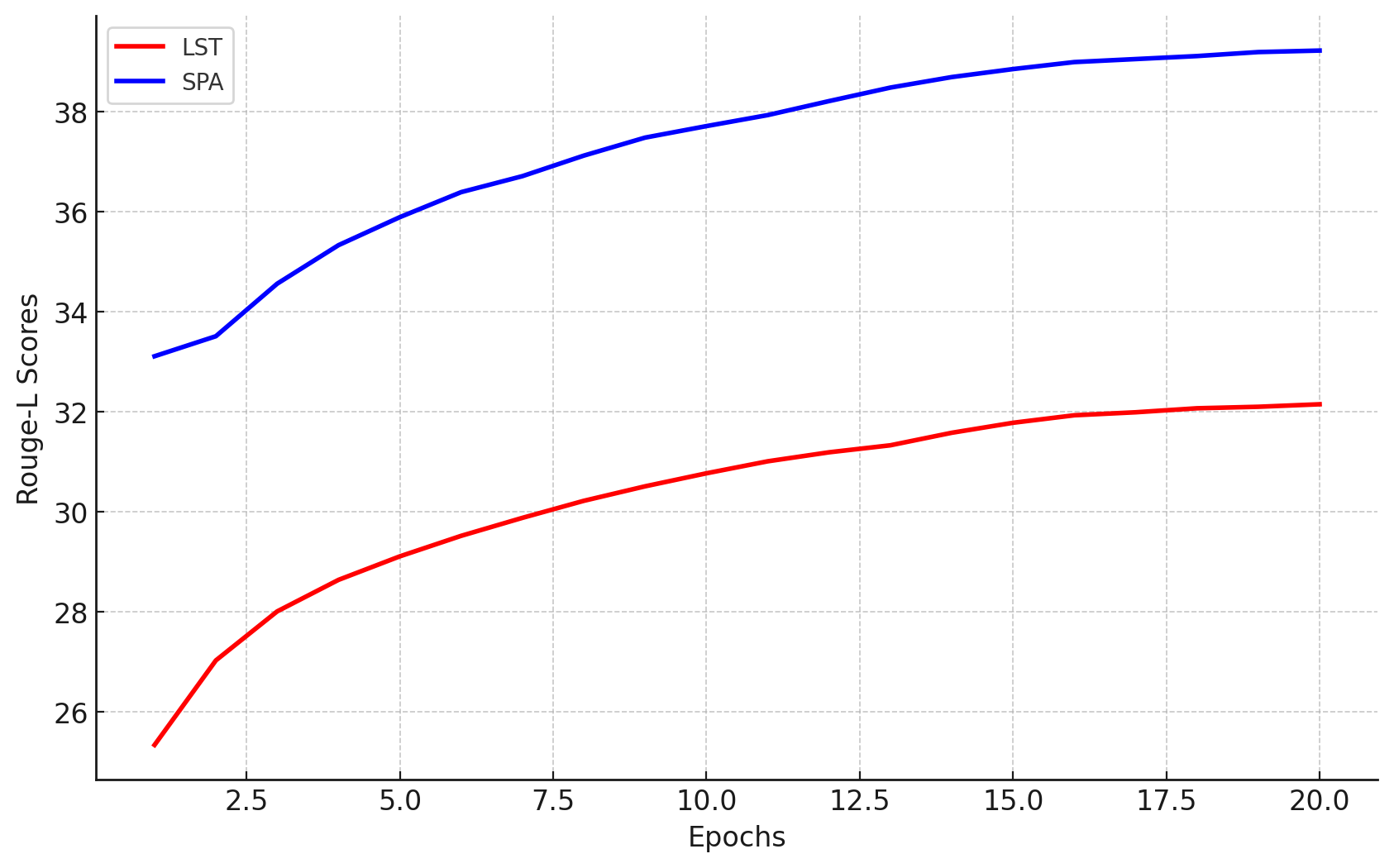}
    \caption{Performance on XSum datasets between LST and SPA.}
    \label{fig:eval}
  \end{minipage}%
  \hfill
   \begin{minipage}[b]{0.46\textwidth}
    \centering
    \includegraphics[width=0.98\linewidth]{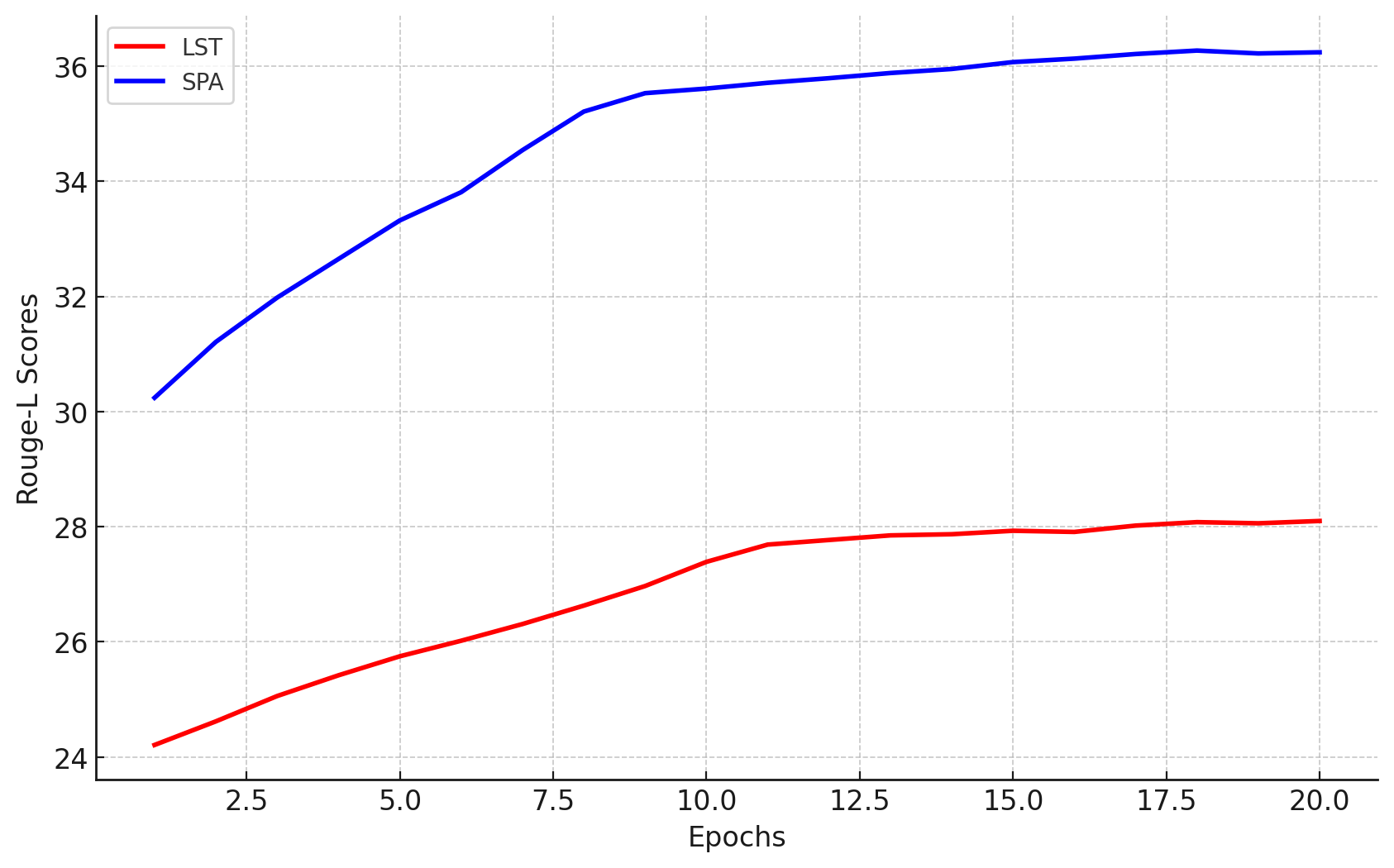}
    \caption{Performance on CNN-DM datasets between LST and SPA.}
    \label{fig:zhu}
  \end{minipage}%
  \hfill
  \vspace{-9pt}
\end{figure*}

\section{Conclusions}
\label{sec:bibtex}

This paper introduces a model suitable for cloud-base and on-devices collaboration. The model leverages the capabilities of LLMs for sequence-to-sequence tasks and utilizes user-specific features from the edge model to guide the overall large model towards our desired objectives. Through a classifier, we can select the most suitable generated sentences. Our approach presents a new paradigm in addressing issues related to edge model deployment, providing valuable insights for future work. We believe that this approach is groundbreaking and can serve as a valuable reference for future research.

%
%
%
%
{
\bibliographystyle{splncs04}
\bibliography{custom}

\begin{thebibliography}{10}
\providecommand{\url}[1]{\texttt{#1}}
\providecommand{\urlprefix}{URL }
\providecommand{\doi}[1]{https://doi.org/#1}

\bibitem{alizadeh2023llm}
Alizadeh, K., Mirzadeh, I., Belenko, D., Khatamifard, K., Cho, M., Del~Mundo, C.C., Rastegari, M., Farajtabar, M.: Llm in a flash: Efficient large language model inference with limited memory. arXiv preprint arXiv:2312.11514  (2023)

\bibitem{bang2023multitask}
Bang, Y., Cahyawijaya, S., Lee, N., Dai, W., Su, D., Wilie, B., Lovenia, H., Ji, Z., Yu, T., Chung, W., et~al.: A multitask, multilingual, multimodal evaluation of chatgpt on reasoning, hallucination, and interactivity. In: Proceedings of the 13th International Joint Conference on Natural Language Processing and the 3rd Conference of the Asia-Pacific Chapter of the Association for Computational Linguistics (Volume 1: Long Papers). pp. 675--718 (2023)

\bibitem{bo2024rootcauseattributiondelivery}
Bo, S., Xiao, M.: Root cause attribution of delivery risks via causal discovery with reinforcement learning (2024), \url{https://arxiv.org/abs/2408.05860}

\bibitem{bo2024attention}
Bo, S., Zhang, Y., Huang, J., Liu, S., Chen, Z., Li, Z.: Attention mechanism and context modeling system for text mining machine translation. arXiv preprint arXiv:2408.04216  (2024)

\bibitem{bolya2022token}
Bolya, D., Fu, C.Y., Dai, X., Zhang, P., Feichtenhofer, C., Hoffman, J.: Token merging: Your vit but faster. In: The Eleventh International Conference on Learning Representations (2022)

\bibitem{brown2020language}
Brown, T., Mann, B., Ryder, N., Subbiah, M., Kaplan, J.D., Dhariwal, P., Neelakantan, A., Shyam, P., Sastry, G., Askell, A., et~al.: Language models are few-shot learners. Advances in neural information processing systems  \textbf{33},  1877--1901 (2020)

\bibitem{cahyawijaya2021greenformers}
Cahyawijaya, S.: Greenformers: Improving computation and memory efficiency in transformer models via low-rank approximation. arXiv preprint arXiv:2108.10808  (2021)

\bibitem{chopra2016abstractive}
Chopra, S., Auli, M., Rush, A.M.: Abstractive sentence summarization with attentive recurrent neural networks. In: Proceedings of the 2016 conference of the North American chapter of the association for computational linguistics: human language technologies. pp. 93--98 (2016)

\bibitem{frantar2023gptq}
Frantar, E., Ashkboos, S., Hoefler, T., Alistarh, D.: Gptq: Accurate post-training quantization for generative pre-trained transformers. In: The Eleventh International Conference on Learning Representations (2023)

\bibitem{gao2016novel}
Gao, H., Wang, H., Feng, Z., Fu, M., Ma, C., Pan, H., Xu, B., Li, N.: A novel texture extraction method for the sedimentary structures’ classification of petroleum imaging logging. In: Pattern Recognition: 7th Chinese Conference, CCPR 2016, Chengdu, China, November 5-7, 2016, Proceedings, Part II 7. pp. 161--172. Springer (2016)

\bibitem{ge-etal-2022-edgeformer}
Ge, T., Chen, S.Q., Wei, F.: {E}dge{F}ormer: A parameter-efficient transformer for on-device seq2seq generation. In: Goldberg, Y., Kozareva, Z., Zhang, Y. (eds.) Proceedings of the 2022 Conference on Empirical Methods in Natural Language Processing. pp. 10786--10798. Association for Computational Linguistics, Abu Dhabi, United Arab Emirates (Dec 2022). \doi{10.18653/v1/2022.emnlp-main.741}, \url{https://aclanthology.org/2022.emnlp-main.741}

\bibitem{gu2024predicting}
Gu, W., Zhong, Y., Li, S., Wei, C., Dong, L., Wang, Z., Yan, C.: Predicting stock prices with finbert-lstm: Integrating news sentiment analysis. arXiv preprint arXiv:2407.16150  (2024)

\bibitem{10021978}
Han, Y., Lee, T.C.M.: Uncertainty quantification for sparse estimation of spectral lines. IEEE Transactions on Signal Processing  \textbf{70},  6243--6256 (2022). \doi{10.1109/TSP.2023.3235662}

\bibitem{10526478}
Han, Y., Lee, T.C.M.: Structural break detection in non-stationary network vector autoregression models. IEEE Transactions on Network Science and Engineering  \textbf{11}(5),  4134--4145 (2024). \doi{10.1109/TNSE.2024.3398002}

\bibitem{houlsby2019parameter}
Houlsby, N., Giurgiu, A., Jastrzebski, S., Morrone, B., De~Laroussilhe, Q., Gesmundo, A., Attariyan, M., Gelly, S.: Parameter-efficient transfer learning for nlp. In: International conference on machine learning. pp. 2790--2799. PMLR (2019)

\bibitem{hu2021lora}
Hu, E.J., Wallis, P., Allen-Zhu, Z., Li, Y., Wang, S., Wang, L., Chen, W., et~al.: Lora: Low-rank adaptation of large language models. In: International Conference on Learning Representations (2021)

\bibitem{hu2023llm}
Hu, Z., Wang, L., Lan, Y., Xu, W., Lim, E.P., Bing, L., Xu, X., Poria, S., Lee, R.: Llm-adapters: An adapter family for parameter-efficient fine-tuning of large language models. In: Proceedings of the 2023 Conference on Empirical Methods in Natural Language Processing. pp. 5254--5276 (2023)

\bibitem{li2021prefix}
Li, X.L., Liang, P.: Prefix-tuning: Optimizing continuous prompts for generation. In: Proceedings of the 59th Annual Meeting of the Association for Computational Linguistics and the 11th International Joint Conference on Natural Language Processing (Volume 1: Long Papers). pp. 4582--4597 (2021)

\bibitem{xinjin2024a}
Li, X., Chang, J., Li, T., Fan, W., Ma, Y., Ni, H.: A vehicle classification method based on machine learning. Preprints  (July 2024). \doi{10.20944/preprints202407.0981.v1}, \url{https://doi.org/10.20944/preprints202407.0981.v1}

\bibitem{xinjin2024intell}
Li, X., Yang, Y., Yuan, Y., Ma, Y., Huang, Y., Ni, H.: Intelligent vehicle classification system based on deep learning and multi-sensor fusion. Preprints  (July 2024). \doi{10.20944/preprints202407.2102.v2}, \url{https://doi.org/10.20944/preprints202407.2102.v2}

\bibitem{li2023loftq}
Li, Y., Yu, Y., Liang, C., Karampatziakis, N., He, P., Chen, W., Zhao, T.: Loftq: Lora-fine-tuning-aware quantization for large language models. In: The Twelfth International Conference on Learning Representations (2023)

\bibitem{lin2020exploring}
Lin, Z., Madotto, A., Fung, P.: Exploring versatile generative language model via parameter-efficient transfer learning. In: Findings of the Association for Computational Linguistics: EMNLP 2020. pp. 441--459 (2020)

\bibitem{ma2024fostc3net}
Ma, D., Li, S., Dang, B., Zang, H., Dong, X.: Fostc3net: A lightweight yolov5 based on the network structure optimization. arXiv preprint arXiv:2403.13703  (2024)

\bibitem{narayan2018don}
Narayan, S., Cohen, S.B., Lapata, M.: Don’t give me the details, just the summary! topic-aware convolutional neural networks for extreme summarization. In: Proceedings of the 2018 Conference on Empirical Methods in Natural Language Processing. pp. 1797--1807 (2018)

\bibitem{openai2023gpt}
OpenAI, R.: Gpt-4 technical report. arXiv pp. 2303--08774 (2023)

\bibitem{qian2020heterogeneous}
Qian, Y., Magginetti, D.J., Jeon, S., Yoon, Y., Olsen, T.L., Wang, M., Gerton, J.M., Yoon, H.P.: Heterogeneous optoelectronic characteristics of si micropillar arrays fabricated by metal-assisted chemical etching. Scientific Reports  \textbf{10}(1),  16349 (2020)

\bibitem{reddy2019coqa}
Reddy, S., Chen, D., Manning, C.D.: Coqa: A conversational question answering challenge. Transactions of the Association for Computational Linguistics  \textbf{7},  249--266 (2019)

\bibitem{reid-etal-2021-subformer-exploring}
Reid, M., Marrese-Taylor, E., Matsuo, Y.: Subformer: Exploring weight sharing for parameter efficiency in generative transformers. In: Moens, M.F., Huang, X., Specia, L., Yih, S.W.t. (eds.) Findings of the Association for Computational Linguistics: EMNLP 2021. pp. 4081--4090. Association for Computational Linguistics, Punta Cana, Dominican Republic (Nov 2021). \doi{10.18653/v1/2021.findings-emnlp.344}, \url{https://aclanthology.org/2021.findings-emnlp.344}

\bibitem{reid2021subformer}
Reid, M., Marrese-Taylor, E., Matsuo, Y.: Subformer: Exploring weight sharing for parameter efficiency in generative transformers. In: Findings of the Association for Computational Linguistics: EMNLP 2021. pp. 4081--4090 (2021)

\bibitem{ren2019collaborative}
Ren, J., Yu, G., He, Y., Li, G.Y.: Collaborative cloud and edge computing for latency minimization. IEEE Transactions on Vehicular Technology  \textbf{68}(5),  5031--5044 (2019)

\bibitem{tao2023sqba}
Tao, Y.: Sqba: sequential query-based blackbox attack. In: Fifth International Conference on Artificial Intelligence and Computer Science (AICS 2023). vol. 12803, p. 128032Q. International Society for Optics and Photonics, SPIE (2017)

\bibitem{tao2023mlad}
Tao, Y.: Meta learning enabled adversarial defense. In: 2023 IEEE International Conference on Sensors, Electronics and Computer Engineering (ICSECE). pp. 1326--1330 (2023). \doi{10.1109/ICSECE58870.2023.10263390}

\bibitem{touvron2023llama}
Touvron, H., Martin, L., Stone, K., Albert, P., Almahairi, A., Babaei, Y., Bashlykov, N., Batra, S., Bhargava, P., Bhosale, S., et~al.: Llama 2: Open foundation and fine-tuned chat models. arXiv preprint arXiv:2307.09288  (2023)

\bibitem{wang2021knowledge}
Wang, L., Yoon, K.J.: Knowledge distillation and student-teacher learning for visual intelligence: A review and new outlooks. IEEE transactions on pattern analysis and machine intelligence  \textbf{44}(6),  3048--3068 (2021)

\bibitem{wang2024research}
Wang, Z., Yan, H., Wang, Y., Xu, Z., Wang, Z., Wu, Z.: Research on autonomous robots navigation based on reinforcement learning. arXiv preprint arXiv:2407.02539  (2024)

\bibitem{welbl2017crowdsourcing}
Welbl, J., Liu, N.F., Gardner, M.: Crowdsourcing multiple choice science questions. In: Proceedings of the 3rd Workshop on Noisy User-generated Text. pp. 94--106 (2017)

\bibitem{xiao2024multiple}
Xiao, M., Bo, S., Wu, Z.: Multiple greedy quasi-newton methods for saddle point problems. arXiv preprint arXiv:2408.00241  (2024)

\bibitem{xu2023llmcad}
Xu, D., Yin, W., Jin, X., Zhang, Y., Wei, S., Xu, M., Liu, X.: Llmcad: Fast and scalable on-device large language model inference (2023)

\bibitem{xu2024applications}
Xu, Q., Feng, Z., Gong, C., Wu, X., Zhao, H., Ye, Z., Li, Z., Wei, C.: Applications of explainable ai in natural language processing. Global Academic Frontiers  \textbf{2}(3),  51--64 (2024)

\bibitem{yang2024research}
Yang, Q., Wang, Z., Liu, S., Li, Z.: Research on improved u-net based remote sensing image segmentation algorithm. arXiv preprint arXiv:2408.11234  (2024), submitted on 22 Aug 2024

\bibitem{zhong2024deep}
Zhong, Y., Liu, Y., Gao, E., Wei, C., Wang, Z., Yan, C.: Deep learning solutions for pneumonia detection: Performance comparison of custom and transfer learning models. medRxiv pp. 2024--06 (2024)

\end{thebibliography}
}

\end{document}